\documentclass[11pt,a4paper]{article}
\usepackage{times,latexsym}
\usepackage{url}
\usepackage[T1]{fontenc}

\usepackage[acceptedWithA]{tacl2018v2}

\usepackage{xspace,mfirstuc,tabulary}

\newif\iftaclinstructions
\taclinstructionsfalse % AUTHORS: do NOT set this to true
\iftaclinstructions
\renewcommand{\confidential}{}
\renewcommand{\anonsubtext}{(No author info supplied here, for consistency with
TACL-submission anonymization requirements)}
\newcommand{\instr}
\fi

\iftaclpubformat % this "if" is set by the choice of options

\else

\fi

\usepackage{amsmath}
\usepackage{amssymb}
\usepackage{graphicx}
\usepackage{caption}
\usepackage{subcaption}
\usepackage{makecell}
\usepackage{multirow}
\usepackage{placeins}

\DeclareMathOperator*{\argmax}{arg\,max}
\DeclareMathOperator*{\relu}{\texttt{ReLU}}

\newcolumntype{?}[1]{!{\vrule width #1}}

\title{No Word is an Island --- a Transformation Weighting Model\\ for Semantic Composition}

\author{
 Corina Dima \and Dani\"{e}l de Kok \and Neele Witte \and Erhard Hinrichs\\
 SFB 833 and Seminar f\"{u}r Sprachwissenschaft, University of T\"{u}bingen \\
 Wilhelmstr. 19, 72074, T\"{u}bingen, Germany \\
 {\fontfamily{pcr}\selectfont
{\small\{corina.dima, daniel.de-kok, neele.witte, erhard.hinrichs\}@uni-tuebingen.de}}\\
}

\date{}

\begin{document}
\maketitle

\setlength{\abovedisplayskip}{5pt}
\setlength{\belowdisplayskip}{5pt}

\begin{abstract}

Composition models of distributional semantics are used to construct
phrase representations from the representations of their
words. Composition models are typically situated on two ends of a
spectrum. They either have a small number of parameters but compose
all phrases in the same way, or they perform word-specific
compositions at the cost of a far larger number of parameters. In this
paper we propose transformation weighting (TransWeight), a composition
model that consistently outperforms existing models on nominal
compounds, adjective-noun phrases and adverb-adjective phrases in
English, German and Dutch. TransWeight drastically reduces the number
of parameters needed compared to the best model in the literature by
composing similar words in the same way.

\end{abstract}

\section{Introduction}
\label{sec:intro}
%count gelb schuh: 13
% schwarz schuh: 197

The phrases \emph{black car} and \emph{purple car} are very
similar --- except for their frequency. In a large corpus, described in
more detail in Section~\ref{sec:datasets}, the phrase \textit{black
car} occurs 131 times --- making it possible to model the whole
phrase distributionally.  But \textit{purple car} is far less
frequent, with only 5 occurrences, and therefore out of reach for
distributional models based on co-occurrence counts.

Distributional word representations derived from large, unannotated
corpora
\cite{collobert2011natural,mikolov2013distributed,pennington2014glove}
capture information about individual words like \textit{purple} and
\textit{car} and are able to express, in vector space, different types
of word similarity (e.g. the similarity between color adjectives like
\textit{black} and \textit{purple}, between \textit{car} and
\textit{truck}, etc.).

Creating phrasal representations, and in particular representing 
low-frequency phrases like \textit{purple car} is a task for
composition models of distributional semantics. A composition model is
a function $f$ that combines the vectors of individual words, e.g. $\mathbf{u}$ for
\textit{black} and $\mathbf{v}$ for \textit{car} into a phrase representation
$\mathbf{p}$.  $\mathbf{p}$ is the result of applying the composition
function $f$ to the word vectors $\mathbf{u}, \mathbf{v}$: $\mathbf{p}
= f(\mathbf{u}, \mathbf{v})$. 

Our goal is to find the function $f$ that learns how to compose phrase
representations by training on a set of phrases. The target
representation of the whole phrase, $\mathbf{\tilde{p}}$, is learned
directly from the corpus, if needed by concatenating first the word
pairs of interest, e.g. \textit{black\_car}. The function $f$ seeks
to maximize the cosine similarity of the composed representation
$\mathbf{p}$ and the target representation $\mathbf{\tilde{p}}$, $
\argmax\frac{\mathbf{p}\cdot\mathbf{\tilde{p}}}{\lVert\mathbf{p}\rVert_2\lVert\mathbf{\tilde{p}}\rVert_2}$.

\vspace{0.5em}
The contributions of this paper are as follows: 
\begin{itemize}
\item We provide an extensive review and evaluation of existing
composition models. The evaluation is carried out in parallel on three
syntactic constructions using English, German, and Dutch treebanks:
adjective-noun phrases (\textit{black car}, \textit{schwarz Auto},
\textit{zwart auto}), nominal compounds (\textit{apple tree},~
\textit{Apfelbaum}, \textit{appelboom}), and adverb-adjective phrases
(\textit{very large}, \emph{sehr gro{\ss}}, \emph{zeer groot}). These
constructions serve as case studies in analyzing the generalization
potential of composition models.

The evaluation of existing approaches shows that state-of-the-art
composition models lie at opposite ends of the spectrum. On one end,
they use the same matrix transformation \cite{socher2010learning} for
all words, leading to a very general composition. On the other end,
there are composition models that use individual vectors
\cite{dima2015reverse} or matrices \cite{socher2012semantic} for each
dictionary word, leading to word-specific compositions with a
multitude of parameters.  Although the latter perform better, they
require training data for each word. These lexicalized composition
models suffer from the \textit{curse of dimensionality}
\cite{bengio2003neural}: the individual word transformations are
learned exclusively from examples containing those words, and do not
benefit from the information provided by training examples involving
different, but semantically related words. In the example above, the
vectors/matrices for \textit{black} and \textit{purple} would be
trained independently, despite the similarity of the words and their
corresponding vector representations.

\item We propose \textit{transformation weighting} --- a new
composition model where `no word is an island'
\cite{donne1624devotions}.  The model draws on the similarity of the
word vectors to induce similar transformations. Its formulation,
presented in Section~\ref{sec:transform_select}, makes it possible to
tailor the composition function even for examples that were not seen
during training: e.g. even if \textit{purple car} is infrequent, the
composition can still produce a suitable representation based on other
phrases containing color adjectives that occur in training.

\item We correct an error in the rank
evaluation methodology proposed by \citet{baroni2010nouns} and subsequently
used as an evaluation standard in other publications
\cite{dinu2013general,dima2015reverse}. The corrected methodology, described
in Section~\ref{sec:evaluation_methodology}, uses the original phrase
representations as a reference point instead of the composed
representations. It makes a fair comparison between the results
of different composition models possible.

\item We provide reference TensorFlow \cite{tensorflow2015-whitepaper}
implementations of all the composition models investigated in this
paper\footnote{\url{https://github.com/sfb833-a3/commix}, last accessed May 16, 2019.} and composition datasets for English and German compounds, adjective-noun and
adverb-adjective phrases and for Dutch adjective-noun and
adverb-adjective phrases\footnote{\url{http://hdl.handle.net/11022/0000-0007-D3BF-4}, last accessed May 16, 2019.}.
\end{itemize}

\section{Previous work in composition models}
\label{sec:previous-work}

Word representations are widely used in modern-day natural language
processing. They improve the lexical coverage of NLP systems by
extrapolating information about words seen in training to semantically
similar words that were not part of the training data. Another
advantage is that word representations can be trained on large, unannotated corpora using
unsupervised techniques based on word co-occurrences. However, the same
modeling paradigm cannot be readily used to model phrases. Due to the
productivity of phrase construction, only a small fraction of
all grammatically-correct phrases will actually occur in corpora.

Composition models attempt to solve this problem through a bottom-up
approach, where a phrase representation is constructed from its parts.
Composition models have succeeded in building representations for English
adjective-noun phrases like \textit{red car} \cite{baroni2010nouns}, nominal compounds
in English --- \textit{telephone number} \cite{mitchell2010composition} and
German --- \textit{Apfelbaum} `apple tree' \cite{dima2015reverse}, determiner
phrases like \textit{no memory} \cite{dinu2013general} and for modeling
derivational morphology in English, e.g.
\textit{re-}+\textit{build}$\rightarrow$\textit{rebuild}
\cite{lazaridou2013compositional} and German, e.g. \textit{taub}
`deaf'+\textit{-heit} $\rightarrow$ \textit{Taubheit} `deafness'
\cite{pado2016predictability}.

\begin{table*}[htb!]

\begin{center}
\begin{tabular}{l|l}
\bf Name        & \bf Composition function $f$ \\
\hline
Addition \cite{mitchell2010composition} & $\mathbf{u}+\mathbf{v}$ \\
SAddition \cite{mitchell2010composition} & $\alpha\mathbf{u}+\beta\mathbf{v}$ \\
VAddition & $\mathbf{a} \odot \mathbf{u}+\mathbf{b} \odot \mathbf{v}$ \\
% a_1, a_2; r, t
Matrix \cite{socher2010learning} & $g(\mathbf{W}[\mathbf{u}; \mathbf{v}] + \mathbf{b})$ \\
FullLex \cite{socher2012semantic} & $g(\mathbf{W}[\mathbf{A}^v \mathbf{u}; \mathbf{A}^u \mathbf{v}] + \mathbf{b})$ \\
BiLinear \cite{socher2013reasoning, socher2013recursive} & $g(\mathbf{u}^\intercal\mathbf{E}\mathbf{v} + \mathbf{W}[\mathbf{u}; \mathbf{v}] + \mathbf{b})$ \\
WMask \cite{dima2015reverse} & $g(\mathbf{W}[\mathbf{u}\odot\mathbf{u^m}; \mathbf{v}\odot\mathbf{v^h}] + \mathbf{b})$\\
\hline
\end{tabular}
\end{center}
\caption{\label{tbl:composition-functions} Composition functions from the literature. $\mathbf{u}$ and $\mathbf{v}$ are the word representations of the inputs. The results reported in Section~\ref{sec:results} use the identity function $g(x)=x$ instead of a nonlinearity because no improvements were observed for these models using $g=\tanh$ or $g=\relu$.}
\end{table*}

In this section, we discuss several composition models from the
literature, summarized in Table~\ref{tbl:composition-functions}. They range
from simple additive models to multi-layered word-specific models. In
our descriptions, the inputs to the composition functions are two vectors,
$\mathbf{u}, \mathbf{v} \in \mathbb{R}^{n}$, where $n$ is the dimensionality
of the word representation. $\mathbf{u}$ and $\mathbf{v}$ are the representations of the first and second element of the phrase and are fixed during
the training process. In this work, the composed representation has the same
dimensionality as the inputs, $\mathbf{p} \in \mathbb{R}^{n}$. This allows us to train the composition functions such that the composed representations are in the same vector space as the word representations.

\paragraph{Additive}

Some of the earliest proposed models were additive models of the form
$\mathbf{u} + \mathbf{v}$ \cite{mitchell2010composition}. The
intuition behind the additive models is that $\mathbf{p}$ lies between
$\mathbf{u}$ and $\mathbf{v}$. The downside of this simple approach is
that it is not sensitive to word order: it would produce the same
representation for \emph{car factory} and \emph{factory car}. This
suggests that $\mathbf{u}$ and $\mathbf{v}$ should be weighted
differently, for example using scaling:
$\alpha\mathbf{u}+\beta\mathbf{v}$ \cite{mitchell2010composition}
or component-wise scaling: $\mathbf{a} \odot \mathbf{u}+\mathbf{b}
\odot \mathbf{v}$.

\paragraph{Matrix} The Matrix model proposed by \citet{socher2010learning}
performs an affine transformation of the concatenated input vectors,
$[\mathbf{u};\mathbf{v}] \in \mathbb{R}^{2n}$, using a matrix
$\mathbf{W}\in \mathbb{R}^{n\times 2n}$ and bias $\mathbf{b}\in
\mathbb{R}^n$.

While the Matrix model is more powerful than the additive models, it
transforms all possible $\mathbf{u}, \mathbf{v}$ pairs in the same
manner. This `one size fits all' approach is counterintuitive: one
would expect, for example, that color adjectives modify nouns in a
different way than adjectives describing a physical quality
(\textit{black car} versus \textit{fast car}). A single transformation
is bound to model a general composition that works reasonably well for
most training examples, but cannot capture more specific word
interactions.

\paragraph{FullLex} The Full\-Lex model\footnote{The name \citet{socher2012semantic}
propose for their model is MV-RNN. However, we found \citet{dinu2013general}'s
renaming proposal, FullLex, to be more descriptive.} proposed by \citet{socher2012semantic}
is a combination of the Matrix model with \citet{baroni2010nouns}'s adjective-specific
linear map model.
FullLex captures word-specific interactions using a trainable tensor
$\mathbf{A} \in \mathbb{R}^{|V| \times n \times n}$, where $|V|$ is the
vocabulary size. $\mathbf{u}$ and $\mathbf{v}$ are transformed crosswise
using the transformation matrices $\mathbf{A}^u, \mathbf{A}^v \in
\mathbb{R}^{n \times n}$. The transformed representations $\mathbf{A}^v
\mathbf{u}$ and $\mathbf{A}^u \mathbf{v}$ are then the input of the Matrix
model. Every matrix $\mathbf{A}^{w}$ in the FullLex model is initialized
using the identity matrix $\mathbf{I}$ with some small perturbations. Since
$\mathbf{I}\mathbf{u} = \mathbf{u}$, the FullLex model starts as an
approximation of the Matrix model. The matrices of the words that occur
in the training data are updated during parameter estimation to better
predict phrase representations.

The FullLex model suffers from two deficiencies, caused by treating each
word as an island: (1)~The FullLex model only learns specialized matrices
for words that were seen in the training data. Words not in the training
data are transformed using the identity matrix, thereby effectively reducing
the FullLex model to the Matrix model for unknown words. (2) Since the model
stores word-specific matrices, the FullLex model has an excessively large
number of parameters, even for modest vocabularies. For example, in our
experiments with English adjective-noun phrases the FullLex model used $\sim$740M
parameters for a vocabulary of 18,481 words. The large number of parameters
makes the model sensitive to overfitting.

The size of the FullLex model can be reduced by using
low-rank matrix approximations for $\mathbf{A}^{w}$
\cite{socher2012semantic}. However, this variation only addresses the model
size problem, it does not improve the handling of unknown words.

\paragraph{BiLinear} The BiLinear model was proposed by
\citet{socher2013reasoning,socher2013recursive} as an alternative to the
FullLex model. This model allows for stronger interactions between word
representations than the Matrix model. At the same time, BiLinear has
fewer parameters when compared to FullLex, because it avoids per-word
matrices. \citet{socher2013reasoning,socher2013recursive}'s goal is to build a
model that is better able to generalize over
different inputs. The core of the bilinear composition model is a tensor
$\mathbf{E} \in \mathbb{R}^{n \times d \times n}$ that stores $d$ bilinear
forms.  Each of the bilinear forms is multiplied by $\mathbf{u}$ and
$\mathbf{v}$ ($\mathbf{u}^\intercal\mathbf{E}\mathbf{v}$) to form a composed
vector of dimensionality $d$. Since the size of the phrase representation is
$n$, in this paper we assume $d = n$. Each bilinear form can be seen as
capturing a different interaction between $\mathbf{u}$ and $\mathbf{v}$. The
model then adds the vector representation computed using the bilinear forms
to the output of a Matrix model.

The BiLinear model solves both problems of the FullLex model. It does not
apply a fallback transformation for every unseen word, while at the same
time drastically reducing the number of parameters. Unfortunately, in our
composition experiments the BiLinear model fares generally worse than
the FullLex model. Consequently, a strong argument can be made in favor of
a model that learns information about specific words or groups of words.

\paragraph{WMask} Another alternative that trains word-specific
transformations but uses fewer parameters than FullLex is the WMask model
proposed by \citet{dima2015reverse}. The reduction in the number of
parameters is achieved by training, for each word, only two mask vectors,
$\mathbf{u}^{m}$, $\mathbf{u}^{h} \in \mathbb{R}^n$, instead of an $n\times n$ matrix. The mask
vectors are positional: if $\mathbf{u}$ is the vector for \textit{leaf}, the
first mask $\mathbf{u}^{m}$ is used to represent \textit{leaf} in phrases
where \textit{leaf} is the first word, i.e. \textit{leaf blower}, while in \textit{autumn
leaf}, where \textit{leaf} is the second word, the second mask, $\mathbf{u}^{h}$, is used.
The trainable parameters are stored in two matrices $\mathbf{W}^m$,
$\mathbf{W}^h$ $\in \mathbb{R}^{|V|\times n}$, where $|V|$ is the size of
the vocabulary.

The mask vectors $\mathbf{u}^{m}$ and $\mathbf{u}^{h}$ are initialized
using vectors of ones, $\mathbf{1}$ and allow the initial word
representations to be fine-tuned for each individual composition. For
words not in the training data WMask is reduced, like FullLex, to the
Matrix model. In contrast to the FullLex model, which uses crosswise
transformations of the input vectors, $\mathbf{A}^v\mathbf{u}$ and
$\mathbf{A}^u\mathbf{v}$, \citet{dima2015reverse} uses direct
transformations: $\mathbf{u}$, the vector of the first word, is
transformed via element-wise multiplication with it's first position
mask, $\mathbf{u}^m$, $\mathbf{u}\odot\mathbf{u}^{m}$. The vector of
the second word, $\mathbf{v}$, is similarly transformed, this time
using the second position mask, $\mathbf{v}\odot\mathbf{v}^{h}$. The
transformed vectors are then composed using the Matrix model. We have
experimented with both direct and crosswise application of masks. The
direct application of masks, as proposed by \citet{dima2015reverse},
provided consistently better results than the crosswise application.

% see VFullLex results - commented out

Although WMask uses fewer parameters, it still has a linear dependence to
the number of words in the vocabulary, $|V|$. As in FullLex's case, the
masks only improve the composition of words seen during training and provide
no benefit for similar words that were not in training.

\paragraph{Non-linearities} Most of the models described in this section can be
used with a non-linear activation function such as the hyperbolic tangent or
$\relu$~\cite{hahnloser2000digital}.

We have experimented with these non-linearities: models performed far worse with $g=\relu$, 
and there was no tangible improvement in model performance using $g=\tanh$. 

We conjecture that a non-linearity is unnecessary because in our
experiments the source and the target representations come from a
vector space that was trained to capture linear substructures
\cite{mikolov2013distributed}. The $\relu$ has the additional
disadvantage that it cannot produce negative vector components in the
target representation. The results reported in
Section~\ref{sec:results} use the identity function, $g(x)=x$, for all
the composition models in Table~\ref{tbl:composition-functions}.

\paragraph{Summary of the state of the art}

Even though the BiLinear and WMask models attempt to address the
shortcomings of the FullLex model, we will show in Section~\ref{sec:results}
that FullLex is the best performer among the models summarized in this
section. However, two crucial shortcomings of the FullLex still need to be
addressed: (1) The FullLex model only learns specialized matrices for words
that were seen in the training data; (2) the FullLex model has an
excessively large number of parameters. In the next section, we will propose
a new composition model that effectively addresses both
shortcomings and outperforms the FullLex model.

\section{The transformation weighting model}
\label{sec:transform_select}

The premise of our model is that words with similar meanings compose with
other words in a similar fashion. In theoretical linguistics, this insight
has been captured by the notion of selectional restrictions. E.g. color
adjectives, such as \emph{black}, \emph{green} and \emph{purple} combine
with concrete objects, such as \emph{apple}, \emph{bike}, and \emph{car}.
A FullLex model composes \emph{black}, \emph{green}, and \emph{purple} with
the nominal head \emph{car} with three different composition functions,
treating each word as an island:
\vspace{0.5em}
\begin{align}
\label{eq:red_green_yellow}
\mathbf{p}^{black\_car} &= \mathbf{W}[\mathbf{A}^{car} \mathbf{u}; \mathbf{A}^{black} \mathbf{v}]
  + \mathbf{b} \\
\mathbf{p}^{green\_car} &=\mathbf{W}[\mathbf{A}^{car} \mathbf{u}; \mathbf{A}^{green} \mathbf{v}]
  + \mathbf{b} \nonumber \\
\mathbf{p}^{purple\_car} &=\mathbf{W}[\mathbf{A}^{car} \mathbf{u}; \mathbf{A}^{purple} \mathbf{v}]
  + \mathbf{b} \nonumber
\end{align}

Because each adjective is transformed via a se\-parate matrix -- $\mathbf{A}^{black}, \mathbf{A}^{green}$ and $\mathbf{A}^{purple}$ in Eq.~\ref{eq:red_green_yellow} -- such a model fails to account for the similarity of the lexical meaning of
color adjectives. The same lack of generalization holds for the lexical
meaning of concrete objects since their composition matrices are also
completely independent of each other. A more appropriate model should be able
to generalize over the lexical meanings of the words that enter phrasal
composition.

Our proposed model, \emph{transformation weighting}, addresses this issue by performing phrasal composition in two stages --- a \emph{transformation} stage and a \emph{weighting} stage. In the transformation stage the model diversifies its treatment of the inputs by applying multiple different transformation matrices to the same input vectors. Because the number of transformations is much smaller than the number of words in the vocabulary, the model is encouraged to reuse transformations for similar inputs. The result of the transformation step is $\mathbf{H} \in \mathbb{R}^{t\times n}$, a set of  $t$ $n$-dimensional combined representations. Each row of $\mathbf{H}$, $\mathbf{H}_i$, is a combination the input vectors $\mathbf{u}$ and $\mathbf{v}$ parametrized by the $i$-th transformation matrix.

In the second stage of the composition process $\mathbf{H}_1, \mathbf{H}_2, ..., \mathbf{H}_t$ are combined into a final composed representation $\mathbf{p}$. We have experimented with several variants for combining the $t$ transformations into a single vector.

\subsection{Applying transformations}
\label{sec:applying_transformations}

Our proposal takes a middle ground between applying the same transformation to each $\mathbf{u}, \mathbf{v}$ --- as in the Matrix case, and applying word-specific transformations from a set of $|V|$ transformations --- as in the FullLex case. $t$, the number of transformations, is a hyperparameter of the model. 
Setting $t=100$ transformations was empirically found to provide the best balance between model size and accuracy. 
We experimented with using between 20 to 500 transformations. Using fewer than 80 transformations resulted in suboptimal performance on all datasets. However, increasing the number of transformations above 100 only provided diminishing returns at the cost of a larger number of parameters.

The transformations are specified via a set of matrices $\mathbf{T}
= [\mathbf{T}^u$; $\mathbf{T}^v] \in \mathbb{R}^{t\times n\times 2n}$, and
the corresponding biases $\mathbf{B} \in
\mathbb{R}^{t\times n}$, where $n$ is the length of the word representation. 
We use Eq.~\ref{eq:hidden_layer_repr} to obtain $\mathbf{H} \in \mathbb{R}^{t \times
n}$, where $g=\relu$
\cite{hahnloser2000digital}.
% We apply $g=\relu$
% \cite{hahnloser2000digital} to obtain $\mathbf{H} \in \mathbb{R}^{t \times
% n}$ in Eq.~\ref{eq:hidden_layer_repr}. 
\vspace{0.3em}
\begin{align}
  \label{eq:hidden_layer_repr}
  \mathbf{H} =~ g(\mathbf{T} [\mathbf{u}; \mathbf{v}] + \mathbf{B})
\end{align}
\vspace{0em}
The next step is to combine the information in $\mathbf{H}$ into a single output representation $\mathbf{p}$, by weighting in the individual contribution of each of the transformations.

\subsection{Weighting}
\label{sec:weighting}

We experimented with four different ways of combining the $t$ rows of $\mathbf{H}$ into $\mathbf{p}$. A first variation, TransWeight-feat, uses a weight vector $\mathbf{w}^{feat} \in \mathbb{R}^n$ and a bias vector $\mathbf{b}^{feat} \in \mathbb{R}^n$ to weight the individual features of each of the $t$ transformed vectors. The weighted vectors are then summed. Each component of $\mathbf{p}$ is obtained via   $\mathbf{p}_c = \mathbf{w}^{feat}_c \bigg[\sum\limits_{i=1}^{t}\mathbf{H}_{i,c}\bigg] + \mathbf{b}^{feat}_c$.

Another weighting variation, TransWeight-trans, uses a weight vector $\mathbf{w}^{trans} \in \mathbb{R}^t$ and a bias $\mathbf{b}^{trans} \in \mathbb{R}^n$ to weight the $t$ transformed vectors. Each $\mathbf{p}_c$ is a weighted sum of the corresponding column of $\mathbf{H}$, $\mathbf{p}_c = \bigg[\sum\limits_{i=1}^{t}\mathbf{H}_{i,c}\mathbf{w}^{trans}_i\bigg] + \mathbf{b}^{trans}_c$.

A third variation, TransWeight-mat, weights the elements of $\mathbf{H}$ using a matrix $\mathbf{W}^{mat} \in \mathbb{R}^{t\times n}$ and the bias $\mathbf{b}^{mat} \in \mathbb{R}^n$. The result of the Hadamard pro\-duct $\mathbf{W}^{mat}\odot\mathbf{H}$ is summed column-wise, resulting in a vector whose components are given by $\mathbf{p}_c = \bigg[\sum\limits_{i=1}^t(\mathbf{W}^{mat}\odot\mathbf{H})_{i,c}\bigg] + \mathbf{b}^{mat}_c$.

Although distinct, the three weighting procedures have a common bias: they perform a \emph{local weighting} of the $t$ rows of $\mathbf{H}$. The local weighting means that the $c$-th component of the final composed representation, $\mathbf{p}_c$, is based only on the values in the $c$-th column of $\mathbf{H}$, and does not integrate information from the other $n-1$ columns. As the results in Table~\ref{tbl:weighting_variations} will show, local weightings are unable to tap into the additional information in $\mathbf{H}$. 
 
In transformation weighting, TransWeight, we use a \emph{global weighting} tensor $\mathbf{W} \in \mathbb{R}^{t \times n \times n}$ and a bias $\mathbf{b} \in \mathbb{R}^n$ to combine the elements of $\mathbf{H}$. The weighting is performed using a tensor double contraction ($:$), as shown in Eq.~\ref{eq:double-contraction}. 
\vspace{0.3em}
\begin{equation}
  \label{eq:double-contraction}
  \mathbf{p} = \mathbf{W} : \mathbf{H} + \mathbf{b}
\end{equation}
\vspace{0em}
In this formulation the $c$-th component of the final representation is obtained as $\mathbf{p}_c=\sum\limits_{i=1}^n\sum\limits_{j=1}^t \mathbf{W}_{c,i,j}\mathbf{H}_{j,i}$. The double dot product operation results in a global weighting because the entire matrix of transformations $\mathbf{H}$ is taken into account for each component of $\mathbf{p}$, albeit using a component-specific weighting. 
% TransWeight takes into account all the different transformations and performs a global weighting when computing each element of the phrasal representation $\mathbf{p}$.

TransWeight addresses the shortcomings of existing composition models identified in Section~\ref{sec:previous-work}. Because the transformation matrices are not word-specific, the number of necessary parameters is drastically reduced. Moreover, learning to reuse transformations for words with similar vector representations is an integral part of training the model. This makes TransWeight particularly adept for creating composed representations of phrases that contain new words. As long as the new words are similar to some of the words seen during training, the model can reuse the learned transformations for building new phrasal representations.

\subsection{Why is the non-linearity necessary for transformation weighting?}

As mentioned earlier, the models described in
Section~\ref{sec:previous-work} do not benefit from the addition of
non-linear activations. This poses the question why a non-linearity is
necessary in the transformation weighting model. When the non-linearity
and biases are omitted, the transformation weighting model is:
\begin{align}
  \mathbf{p} =~ & \mathbf{W} : (\mathbf{T} [\mathbf{u}; \mathbf{v}])
  \nonumber \\
  \mathbf{p}_c =~
    & \sum_{i=1}^n \sum_{j=1}^t \left[
      \mathbf{W}_{c,i,j}\left(\mathbf{T}[\mathbf{u};\mathbf{v}]\right)_{j,i}
        \right] \nonumber \\
      & \sum_{i=1}^n \sum_{j=1}^t \left[
        \mathbf{W}_{c,i,j}\left(\sum_{k=1}^{2n} \mathbf{T}_{j,i,k}
      [\mathbf{u};\mathbf{v}]_k\right) \right]
         %%& \sum_i \sum_j \sum_k \mathbf{W}_{i,j,c} \mathbf{T}_{i,j,k}
         %%[\mathbf{u};\mathbf{v}]_k
\end{align}

\noindent $\mathbf{W}_{c,*,*}$ is a matrix of component-specific
weightings of the transformed representations $\mathbf{T}[\mathbf{u};\mathbf{v}]$.
We replace $\mathbf{T}$ by a component-specific transformation tensor
$\mathbf{T}^c = \mathbf{W}_{c,*,*} \odot \mathbf{T}$: 
% \vspace{0.5em}
\begin{align}
  \mathbf{p}_c =~ & \sum_{i=1}^n \sum_{j=1}^t \sum_{k=1}^{2n} \mathbf{T}^c_{j,i,k} [\mathbf{u};\mathbf{v}]_k
  \nonumber \\
    =~ & \sum_{i=1}^n \sum_{j=1}^t \mathbf{T}^c_{j,i} \cdot [\mathbf{u};\mathbf{v}]
\end{align}

\noindent If $\mathbf{t}^c = \sum_i \sum_j \mathbf{T}^c_{j,i}$, then it
follows from the distributive property of the dot product that $\mathbf{p}_c
= \mathbf{t}^c \cdot [\mathbf{u};\mathbf{v}]$. Without a non-linearity, the
transformation weighting model thus reduces to the Matrix model. This does
not hold for the transformation with the non-linearity --- since
$\alpha g(a) \neq g(\alpha a)$ for a non-linearity $g$, we cannot precompute
a component-specific transformation matrix $\mathbf{T}^c$.

\section{Training and evaluating composition models}
\label{sec:evaluating_composition_models}

We evaluated the composition models described in this paper on three
phrase types: compounds; adjective-noun phrases; and adverb-adjective
phrases for three languages: English; German; and Dutch. As discussed
in Section~\ref{sec:intro}, our goal is to train and evaluate
composition functions of the form $\mathbf{p} =
f(\mathbf{u},\mathbf{v})$, such that the cosine similarity between the
predicted representation $\mathbf{p}$ and the target representation
$\tilde{\mathbf{p}}$ is maximized. In order to do so, we need a target
representation $\tilde{\mathbf{p}}$ for each phrase as well as the
representations of their constituent words, $\mathbf{u}$ and
$\mathbf{v}$.

In Section~\ref{sec:datasets}, we describe the treebanks that were used
to train the word and phrase representations $\mathbf{u}$,
$\mathbf{v}$, and
$\tilde{\mathbf{p}}$. Section~\ref{sec:phrase-extraction} describes
how the phrase sets were obtained for each language. In
Section~\ref{sec:training}, we describe how the word/phrase
representations and composition models were trained. Finally, in
Section~\ref{sec:evaluation_methodology}, we describe our evaluation
methodology.

\subsection{Treebank}
\label{sec:datasets}

\paragraph{English} The distributional representations for words and
phrases for English were trained on the \textsc{encow16ax} treebank
\cite{SchaeferBildhauer2012full,Schaefer2015b}. \textsc{encow16ax} contains crawled
web data from a wide variety of sources. We extract sentences from
documents with a document quality estimation of \emph{a} or \emph{b}
to obtain a large number of relatively clean sentences. The extracted
subset contains 89.0M sentences and 2.2B tokens. In contrast to Dutch
and German, we train the word representations on word forms, due to
the large number of words that were assigned an \emph{unknown} lemma.

\paragraph{German}

We use three sections of the T\"uBa-D/DP treebank \cite{dekok2019stylebook} to train lemma and
phrase representations for German: (1) articles from the German
newspaper taz from 1986 to 2009; (2) the German Wikipedia dump of
January 20, 2018; and (3) the German proceedings from the EuroParl
corpus \cite{koehn2005europarl,tiedemann2012}. Together, these sections
contain 64.9M sentences and 1.3B tokens.

\paragraph{Dutch}

Lemma and phrase representations for Dutch were trained on the Lassy
Large treebank \cite{van2013large}. The Lassy Large treebank consists
of various genres of written text, such as newspapers, Wikipedia, and text
from the medical domain. The Lassy Large treebank consists of 47.6M
sentences and 700M tokens.

\subsection{Phrase sets}
\label{sec:phrase-extraction}

We extracted the compound sets from existing lexical resources that
are available for English, German, and Dutch. The adjective-noun and
adverb-adjective phrases were automatically extracted from the
treebanks using part-of-speech and dependency annotations. Each phrase
set consists of phrases and their constituents. For example, the
German compound set contains the compound \emph{Apfelbaum} `apple
tree' together with its constituent words \emph{Apfel} `apple` and
\emph{Baum} `tree'. We filter out phrases where either the words or
the phrase do not make the frequency threshold of word2vec
training (Section~\ref{sec:training}). The phrase set sizes are shown in
Table~\ref{tbl:composition-datasets}. Each phrase set was was divided
into \texttt{train}, \texttt{test} and \texttt{dev} splits with the
ratio \emph{7:2:1}.

\begin{table}[htb!]
\scriptsize
\tabcolsep=0.1cm

\begin{center}
\begin{tabular}{l|l|r|r|r|r}
\bf Language        & \bf Phrase Type & \bf Phrases & \bf w1 & \bf w2 & \bf w1 \& w2 \\
\hline
German & Adj-Noun & 119,434 & 7,494 & 16,557 & 24,031\\
& Compounds & 32,246 & 5,661 & 4,899 & 8,079 \\
& Adv-Adj & 23,488 & 1,785 & 5,123 & 5,905\\
\hline
English & Adj-Noun & 238,975 & 8,210 & 13,045 & 20,644\\
& Compounds & 16,978 & 3,689 & 3,476 & 5,408\\
& Adv-Adj & 23,148 & 820 & 3,086 & 3,817\\
\hline
Dutch & Adj-Noun & 83,392 & 4,999 & 10,936 & 15,744\\
& Compounds & 17,773 & 3,604 & 3,495 & 5,317 \\
& Adv-Adj & 4,540 & 476 & 1,050 & 1,335\\
\hline
\end{tabular}
\end{center}
\caption{\label{tbl:composition-datasets} Overview of the composition datasets evaluated in this paper. Reports the total number of phrases (\textbf{Phrases}), the number of unique words in the first (\textbf{w1}) and second (\textbf{w2}) position, as well as the number of unique words in either position (\textbf{w1 \& w2}) for each language, phrase type pair.}
\end{table}

\paragraph{Compounds}

For German, we use the data set introduced by \citet{dima2015reverse}, which was extracted from the German wordnet GermaNet \cite{Hamp:1997,henrich2011determining}.
Dutch noun-noun compounds were extracted from the Celex lexicon
\cite{baayen1993celex}. The English compounds come from the \cite{Tratz:PhD} dataset and from the English WordNet \cite{fellbaum1998wordnet} (the \texttt{data.noun} entries with two constituents separated by dash or underscore).

\paragraph{Adjective-noun phrases}

The adjective-noun phrases were extracted automatically from the
treebanks based on the provided part-of-speech annotations. We treat
every occurrence of an attributive adjective followed by a noun as a
single unit. The adjectives and nouns that are part of such phrases
are therefore absorbed by this unit. The representations of adjectives
and nouns are as a consequence based only on the remaining occurrences
(e.g.\ adjectives in predicative positions and nouns not preceded by
adjectives).

\paragraph{Adverb-adjective phrases}

For the adverb-adjective data sets, we extracted every
\emph{head}-\emph{dependent} pair where: (1) \emph{head} is an
attributive or predicative adjective; and (2) \emph{head} governs
\emph{dependent} with the adverb relation. We did not impose any
requirements with regards to the part of speech of \emph{dependent} in
order to extract both real adverbs (eg. Dutch: \emph{\textbf{zeer}
giftig} `very poisonous') and adjectives which function as an adverb
(eg. Dutch: \emph{\textbf{buitensporig} groot} `exceptionally large').

To be able to learn phrase representations in word2vec's training
regime (Section~\ref{sec:training}), we additionally require that the
dependent immediately precedes the head. Similarly to adjective-noun
phrases, the representations of adverbs and adjectives are
consequently based on the remaining occurrences.

\subsection{Training}
\label{sec:training}

For each phrase type, each target representation $\tilde{\mathbf{p}}$
was trained jointly with the representations of the constituent words
$\mathbf{u}$ and $\mathbf{v}$ using word2vec
\cite{mikolov2013distributed} and the hyperparameters in
Appendix~\ref{appendix:embed-hyperparameters}. For phrases where
words are separated by a space (adjective-noun phrases,
adverb-adjective phrases, and English compounds), we first merged the
phrase into a single unit. This accommodates training using word2vec,
which uses tokens as its basic unit.

Each composition model $\mathbf{p} = f(\mathbf{u},\mathbf{v})$ (with
exception of the unscaled additive model) was trained using backpropagation
with the Adagrad algorithm \cite{duchi2011adaptive}. Since our goal is
to maximize the cosine similarity between the predicted phrase
representation $\mathbf{p}$ and the target representation
$\tilde{\mathbf{p}}$, we used the cosine distance, $1 -
\frac{\mathbf{p}\cdot\mathbf{\tilde{p}}}{\lVert\mathbf{p}\rVert_2\lVert\mathbf{\tilde{p}}\rVert_2}$,
as the loss function. The training hyperparameters are summarized in
Appendix~\ref{appendix:hyperparameters}.

\subsection{Evaluation methodology} 
\label{sec:evaluation_methodology}

\begin{figure}[htbp]
	\centering
	\includegraphics[scale=0.8]{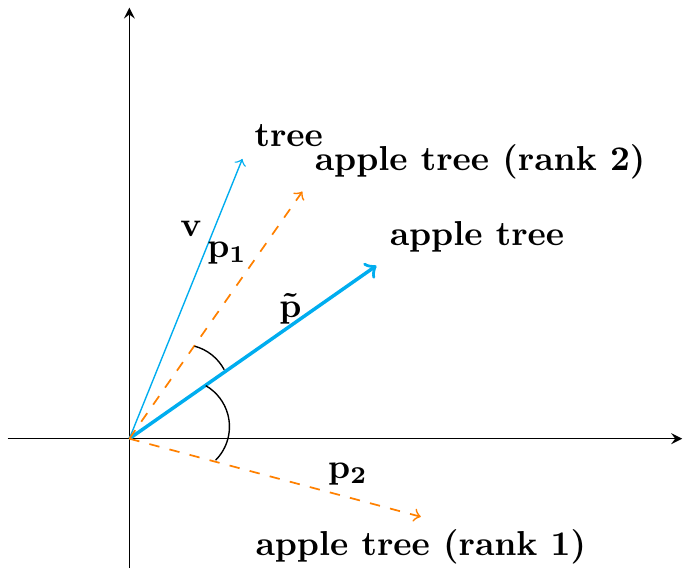}
    \caption{\label{fig:new_rank_evaluation} Corrected evaluation: the original representation $\mathbf{\tilde{p}}$ of the compound \textit{apple tree} is the vector of reference; both original ($\mathbf{v}$) and composed ($\mathbf{p_1}, \mathbf{p_2}$) representations are compared to $\mathbf{\tilde{p}}$ in terms of cosine similarity.}
\end{figure}

\citet{baroni2010nouns} introduced the idea of using a rank evaluation to assess the performance of different composition models. Fig.~\ref{fig:new_rank_evaluation} illustrates the process of evaluating two composed representations, $\mathbf{p_1}$ and $\mathbf{p_2}$, of the compound \textit{apple tree}, produced by two distinct composition functions, $f_1$ and $f_2$. In the simplified setup of Fig.~\ref{fig:new_rank_evaluation}, the original vectors, depicted using solid blue arrows, are $\mathbf{v}$, the vector of \textit{tree} and $\mathbf{\tilde{p}}$, the vector of \textit{apple tree}. $\mathbf{p_1}$ and $\mathbf{p_2}$, the representations composed using $f_1$ and $f_2$, are depicted using dashed orange arrows. 

$\mathbf{p_1}$'s evaluation proceeds as follows: first, $\mathbf{p_1}$ is compared, in terms of cosine similarity, to the original representations of all words and compounds in the dictionary. 
% We measure the cosine of the angle between $\mathbf{p_1}$ and $\mathbf{v}$, and that of the angle between $\mathbf{p_1}$ and $\mathbf{\tilde{p}}$. 
The original vectors are then sorted such that the most similar vectors are first. In Fig.~\ref{fig:new_rank_evaluation}, $\mathbf{v}$, the vector of \textit{tree}, is closer to $\mathbf{p_1}$ than $\mathbf{\tilde{p}}$, the original vector of \textit{apple tree}. The rank assigned to a composed representation is the position of the corresponding original vector in the similarity-sorted list. In $\mathbf{p_1}$'s case, the rank is 2 because the original representation of \textit{apple tree}, $\mathbf{\tilde{p}}$, was second in the ordering.

The same procedure is then performed for $\mathbf{p_2}$. $\mathbf{p_2}$ is compared to the original vectors $\mathbf{\tilde{p}}$ and $\mathbf{v}$ and assigned the rank 1, because $\mathbf{\tilde{p}}$, the original vector for \textit{apple tree}, is its nearest neighbor.

Herein lies the problem: although $\mathbf{p_1}$ is closer to the original representation $\mathbf{\tilde{p}}$ than $\mathbf{p_2}$, $\mathbf{p_1}$'s rank is worse. This is because the vector of reference is the composed representation, which \textit{changes} from one composition function to the other.
\citet{baroni2010nouns}'s formulation of the rank assignment procedure can lead to situations where composed representations rank better even as the distance between the composed and the original vector increases, as illustrated in Fig.~\ref{fig:new_rank_evaluation}.

We propose a simple fix for this issue: we compute all the cosine similarities with respect to $\mathbf{\tilde{p}}$, the original representation of each compound or phrase. Having a fixed reference point makes it possible to correctly assess and compare the performance of different composition models. In the new formulation $\mathbf{p_1}$ is correctly judged to be the better composed representation, assigned rank 1, whereas $\mathbf{p_2}$ is assigned rank 2. 
% (because $\mathbf{v}$ is closer to $\mathbf{\tilde{p}}$ than $\mathbf{p_2}$).

Composition models are evaluated on the \texttt{test} split of each dataset. First, the rank of each composed representation is determined using the procedure described above. Then the ranks of all the \texttt{test} set entries are sorted. We report the first (Q$_1$), second (Q$_2$) and third (Q$_3$) quartile, where Q$_2$ is the median of the sorted rank list, Q$_1$ is the median of the first half and Q$_3$ is the median of the second half of the list. We report two additional performance metrics: the \emph{average cosine distance} (cos-d) between the composed and the original representations of each \texttt{test} set entry and the \emph{percentage of \texttt{test} compounds with rank $\leq$5}. Typically, a composed representation can have close neighbors that are different but semantically similar to the composed phrase. A rank $\leq$5 indicates a well-build representation, compatible with the neighborhood of the original phrase.

% @Corina: I removed this, since it was specific to German.
% The vocabulary
% contains 476,137 words and phrases in the adjective-noun case and
% 403,030 words in the compounds case.

Another particularity of our evaluation is that the ranks are computed
against the full vocabulary of each embedding set.  This is in
contrast to previous evaluations
\cite{baroni2010nouns,dima2015reverse} where the rank was computed
against a restricted dictionary containing only the words and phrases
in the dataset. A restricted dictionary makes the evaluation easier
since many similar words are excluded. In our case similar words from
the entire original vector space can become foils for the composition
model, even if they are not part of the dataset. For example the English 
compounds dataset has a restricted vocabulary of 25,807 words, whereas 
the full vocabulary contains 270,941 words.

% EN Adj-N dataset: 278,505 unique words and phrases in the restricted vocab; the full embedding vocab has 478,373
% EN compounds dataset: 25,807 unique words and phrases in the restricted vocab; full embedding vocab 270,941

\section{Results}
\label{sec:results}

Section~\ref{sec:weighting_variants_results} reports on the performance of the different weighting variants proposed for the TransWeight model on the most challenging of the nine datasets, the German compounds dataset. TransWeight with the best weighting is then compared to existing composition models in Section~\ref{sec:comparison_results}.

\subsection{Performance of different weighting variants}
\label{sec:weighting_variants_results}

Table~\ref{tbl:weighting_variations} compares the performance of the four weighting variants introduced in Section~\ref{sec:weighting}. 

TransWeight-feat, which sums the transformed representations and then weights each component of the summed representation, has the weakest performance,  with only 50.82\% of the test compounds receiving a rank that is lower than 5. 

A better performance -- 52.90\% -- is obtained by applying the same weighting for each column of the transformations matrix $\mathbf{H}$. The results of TransWeight-trans are interesting in two  respects: first, it outperforms the feature variation, TransWeight-feat, despite training a smaller number of parameters (300 vs. 400 in our setup). Second, it performs on par with the TransWeight-mat variation, although the latter has a larger number of parameters (20,200 in our setup). This suggests that an effective combination method needs to take into account full transformations, i.e. entire rows of $\mathbf{H}$ and combine them in a systematic way. 

TransWeight builds on this insight by making each element of the final composed representation $\mathbf{p}$ dependent on each component of the transformed representation $\mathbf{H}$. The result is a noteworthy increase in the quality of the predictions, with $\sim$12\% more of the \texttt{test} representations having a rank $\leq$5.

Although this weighting does use significantly more parameters than the previous weightings (4,000,200 parameters), the number of parameters is relative to the number of transformations $t$ and does not grow with the size of the vocabulary. As the results in the next subsection show, a relatively small number of transformations is sufficient even for larger training vocabularies.

\begin{table}[htb!]
\footnotesize
\tabcolsep=0.08cm
\centering
\begin{tabular}{l|r|r|ccc|r}
\textbf{Model} & \textbf{W param} & \textbf{Cos-d} & \textbf{Q$_1$} & \textbf{Q$_2$} & \textbf{Q$_3$} & \textbf{$\leq$5}   \\
\hline

TransWeight-feat & $n+n$ & 0.344 & 2 & 5 & 28 & 50.82\% \\
TransWeight-trans & $t+n$ & 0.338 & 2 & 5 & 24 & 52.90\% \\
TransWeight-mat & $tn+n$ & 0.338 & 2 & 5 & 25 & 53.24\% \\
\textbf{TransWeight} & $tn^2+n$& \textbf{0.310} & \textbf{1} & \textbf{3} & \textbf{11} & \textbf{65.21\%} \\
% \hline
% TransWeight-r & 100 & 0.4 &  & 0.349 & 1 & 2 & 7 & 71.95\% \\
% TransWeight-c & 100 & 0.6 & & 0.346 & 1 & 2 & 6 & 73.48\% \\
% TransWeight-v & 100 & 0.6 & & 0.345 & 1 & 2 & 6 & 73.69\%\\
% TransWeight & 100 & 0.6 & & \textbf{0.333} & \textbf{1} & \textbf{2} & \textbf{4} & \textbf{78.52\%} \\
% \hline
% TransWeight-r & 100 & 0.2 &  & 0.321 & 1 & 1 & 3 & 82.13\% \\
% TransWeight-c & 100 & 0.6 & & 0.318 & 1 & 1 & 3 & 83.14\% \\
% TransWeight-v & 100 & 0.4 & & 0.318 & 1 & 1 & 3 & 83.12\%\\
% TransWeight & 100 & 0.8 & & \textbf{0.297} & \textbf{1} & \textbf{1} & \textbf{2} & \textbf{89.28\%} \\

\hline
\end{tabular}
\caption{\label{tbl:weighting_variations} Different weighting variations evaluated on the compounds dataset (32,246 nominal compounds). All variations use $t=100$ transformations, word representations with $n=200$ dimensions and the dropout rate that was observed to work best on the \texttt{dev} dataset (see Appendix~\ref{appendix:hyperparameters} for details). Results on the 6442 compounds in the \texttt{test} set of the German compounds dataset.
}

%% Using tf default init; lr=0.01, batch size 100
%% Hyperparameters on the compounds dataset
%% TransWeight-r - dropout 0.4
%% TransWeight-c - dropout 0.6
%% TransWeight-v - dropout 0.6
%% TransWeight - dropout 0.8

%% results on turing: /home/corina/results/tacl_final
\end{table}

\subsection{Comparison to existing composition models}
\label{sec:comparison_results}

\begin{table*}[tbh!]
\footnotesize
\tabcolsep=0.14cm

\begin{tabular}{l?{0.4mm}r|rrr|c?{0.4mm}r|rrr|c?{0.4mm}r|rrr|c}
& \multicolumn{5}{c?{0.4mm}}{\textbf{Nominal Compounds}} & \multicolumn{5}{c?{0.4mm}}{\textbf{Adjective-Noun Phrases}} & \multicolumn{5}{c}{\textbf{Adverb-Adjective Phrases}} \\
\hline
\textbf{Model} & \textbf{Cos-d} & \textbf{Q$_1$} & \textbf{Q$_2$} & \textbf{Q$_3$} & \textbf{$\leq$5}
& \textbf{Cos-d} & \textbf{Q$_1$} & \textbf{Q$_2$} & \textbf{Q$_3$} & \textbf{$\leq$5}
& \textbf{Cos-d} & \textbf{Q$_1$} & \textbf{Q$_2$} & \textbf{Q$_3$} & \textbf{$\leq$5}\\
\hline

\hline
& \multicolumn{15}{c}{\textbf{English}} \\
\hline

Addition  &  0.408 & 2 & 7 & 38 & 46.14\%
& 0.431 & 2 & 7 & 32 & 44.25\%
& 0.447 & 2 & 5 & 15 & 53.01\% \\
SAddition & 0.408 &	2 & 7 & 38 & 46.14\%
& 0.421 & 2 & 5 & 26 & 50.95\%
& 0.420 & 1 & 3 & 8 & 67.76\% \\
VAddition &  0.403 & 2 & 6 & 33 & 47.95\% 
& 0.415 & 2 & 5 & 22 & 53.30\%
& 0.410 & 1 & 2 & 6 & 71.94\% \\
Matrix &  0.354 & 1 & 2 & 9 & 67.37\% 
& 0.365 & 1 & 2 & 6 & 74.38\%
& 0.343 & 1 & 1 & 2 & 91.17\%\\
WMask+ & 0.344 & 1 & 2 & 7 & 71.53\%
& 0.342 & 1 & 1 & 3 & 82.67\%
& 0.335 & 1 & 1 & 2 & 93.27\% \\
BiLinear & 0.335 & 1 & 2 & 6 & 73.63\% 
& 0.332 & 1 & 1 & 3 & 85.32\%
& 0.331 & 1 & 1 & 1 & 93.59\%\\ 
FullLex+ & 0.338 & 1 & 2 & 7 & 72.82\%
& 0.309 & 1 & 1 & 2 & 90.74\%
& 0.327 & 1 & 1 & 1 & 94.28\%\\
\textbf{TransWeight} & \textbf{0.323} & \textbf{1} & \textbf{1} & \textbf{4.5} & \textbf{77.31\%} 
& \textbf{0.307} & \textbf{1} & \textbf{1} & \textbf{2} & \textbf{91.39\%}
& \textbf{0.311} & \textbf{1} & \textbf{1} & \textbf{1} & \textbf{95.78\%} \\

\hline
& \multicolumn{15}{c}{\textbf{German}} \\
\hline

Addition  & 0.439 & 9 & 48 & 363 & 17.49\%
& 0.428 & 4 & 13 & 71 & 32.95\%
& 0.500 & 4 & 19 & 215.5 & 29.87\% \\
SAddition & 0.438 & 9 & 46 & 347 & 18.02\%
& 0.414 & 2 & 8 & 53 & 42.80\%
& 0.473 & 2 & 7 & 99.5 & 45.44\% \\
VAddition & 0.430 & 8 & 39 & 273 & 19.02\%
& 0.408 & 2 & 7 & 43 & 45.14\% 
& 0.461 & 2 & 5 & 52 & 51.12\% \\
Matrix &  0.363 & 3 & 8 & 45 & 41.88\% 
& 0.355 & 1 & 2 & 8 & 68.67\% 
& 0.398 & 1 & 1 & 5 & 76.41\% \\
WMask+ & 0.340 & 2 & 5 & 25 & 52.05\%
& 0.332 & 1 & 2 & 5 & 77.68\%
& 0.387 & 1 & 1 & 3 & 80.94\% \\ 
BiLinear & 0.339 & 2 & 5 & 26 & 53.46\%
& 0.322 & 1 & 1 & 3 & 81.84\%
& 0.383 & 1 & 1 & 3 & 83.02\% \\ 
FullLex+ & 0.329 & 2 & 4 & 20 & 56.83\% 
& 0.306 & 1 & 1 & 2	& 86.29\%
& 0.383 & 1 & 1 & 3 & 83.13\%\\
\textbf{TransWeight} & \textbf{0.310} & \textbf{1} & \textbf{3} & \textbf{11} & \textbf{65.21\%}  
& \textbf{0.297} & \textbf{1} & \textbf{1} & \textbf{2} & \textbf{89.28\%}
& \textbf{0.367} & \textbf{1} & \textbf{1} & \textbf{2} & \textbf{87.17\%} \\

\hline
& \multicolumn{15}{c}{\textbf{Dutch}} \\
\hline

Addition  &  0.477 & 5 & 27 & 223.5 & 27.74\%
& 0.476 & 3 & 13 & 87 & 35.63\%
& 0.532 & 3 & 9 & 75 & 38.04\% \\
SAddition & 0.477 &	5 & 27 & 221 & 27.71\%
& 0.462 & 2 & 7 & 65 & 44.95\%
& 0.503 & 2 & 4 & 34 & 55.57\% \\
VAddition &  0.470 & 4 & 22 & 177 & 29.09\% 
& 0.454 & 2 & 6 & 47 & 48.13\%
& 0.486 & 1 & 3 & 14 & 63.18\%\\
Matrix &  0.411 & 2 & 5 & 26 & 52.19\%
& 0.394 & 1 & 2 & 6 & 74.92\% 
& 0.445 & 1 & 1 & 4 & 78.39\%\\
WMask+ & 0.378 & 1 & 3 & 15 & 60.14\% 
& 0.378 & 1 & 1 & 4 & 80.78\% 
& 0.429 & 1 & 1 & 2 & 83.02\%\\ 
BiLinear & 0.375 & 1 & 3 & 19 & 59.23\% 
& 0.375 & 1 & 1 & 3 & 81.50\% 
& 0.426 & 1 & 1 & 2 & 83.57\% \\ 
FullLex+ & 0.388 & 1 & 3 & 14 & 60.84\%
& 0.362 & 1 & 1 & 2 & 85.24\% 
& 0.433 & 1 & 1 & 3 & 82.36\% \\
\textbf{TransWeight} & \textbf{0.376} & \textbf{1} & \textbf{2} & \textbf{11} & \textbf{66.61\%} 
& \textbf{0.349} & \textbf{1} & \textbf{1} & \textbf{2} & \textbf{88.55\%}
& \textbf{0.423} & \textbf{1} & \textbf{1} & \textbf{2} & \textbf{84.01\%} \\

\hline
\end{tabular}

\caption{\label{tbl:results_all} Results for English, German and Dutch on the composition of nominal compounds, adjective-noun phrases and adverb-adjective phrases.}
\end{table*}

The composition models discussed so far were evaluated on the nine datasets introduced in Section~\ref{sec:datasets}. 
% using three languages -- English, German and Dutch -- and three phrase types -- compounds, adjective-noun and adverb-adjective. 
The results using the corrected rank evaluation procedure described in Section~\ref{sec:evaluation_methodology} are presented in Table~\ref{tbl:results_all}.\footnote{ Table~\ref{tbl:results_oldEval} of the appendix presents for completion the results using the \citet{baroni2010nouns} rank evaluation for the four best performing models. Using the composed representation as a reference point leads to an unfair evaluation of composition models and should be avoided.}

TransWeight, the composition model proposed in this paper, delivers
consistent results, being the best performing model across \textit{all
languages and phrase types}. The difference in performance to the
runner-up model, FullLex+, translates into more of the test phrases
being close to the original representations, i.e. achieving a rank
$\leq5$. This difference ranges from 8\% of the test phrases in the
German compounds dataset to less than 1\% for English adjective-noun
phrases. However, it is important to note the substantial difference
in the number of parameters used by the two models: all TransWeight
models use 100 transformations and have, therefore, a constant number
of 12,020,200 parameters. In contrast the number of parameters used by
FullLex+ increases with the size of the training vocabulary, reaching
739,320,200 parameters in the case of the English adjective-noun
dataset.

The most difficult task for all the composition models in any of the
three languages is compound composition.  We believe this difficulty
can be mainly attributed to the complexity introduced by the
position. For example in adjective-noun composition, the adjective
always takes the first position, and the noun the second. However, in
compounds the same noun can occur in both positions throughout
different training examples. Consider for example the compounds
\textit{boat house} and \textit{house boat}. In \textit{boat house} --
a house to store boats -- the meaning of \textit{house} is shifted
towards \textit{shelter for an inanimate object}, whereas
\textit{house boat} selects from \textit{house} aspects related to
human beings and their daily lives happening on the boat. These
position-related differences can make it more challenging to create
composed representations.

Another aspect that makes the adverb-adjective and adjective-noun datasets easier is the high dataset frequency of some of the adverbs/adjectives.
%For example in the English adjective-noun dataset a small handful of adjectives are extremely frequent, occurring more 
%than 200 times in the training portion of the adjective-noun sample dataset. 7 adjectives (\textit{neu}, \textit{deutsch}, \textit{erst}, \textit{weiter}, \textit{ander}, \textit{eigen}, \textit{klein}) have 686 occurrences in the test data. Because the adjective is always the first element of the composition,
%than 1,000 times in the training portion of the adjective-noun sample dataset. 7 adjectives (\textit{new}, \textit{other}, \textit{more}, \textit{own}, \textit{many}, \textit{such}, \textit{great}) have 6,436 occurrences in the test data. Because the adjective is always the first element of the composition, this amounts to 13.4\% of the \texttt{test} data. Frequent constituents are more likely to be modeled correctly by composition -- thus leading to better results.
For example, in the English adjective-noun dataset a small subset of 52 adjectives like \textit{new}, \textit{good}, \textit{small}, \textit{public}, etc. are extremely frequent, occurring more than 500 times in the training portion of the adjective-noun sample dataset. Because the adjective is always the first element of the composition, the phrases that include these frequent adjectives amount to around 24.8\% of the \texttt{test} dataset. Frequent constituents are more likely to be modeled correctly by composition -- thus leading to better results.
% 13 more than 1000 times in training, 53 more than 500 in training
% 23 more than 200 times in test

% 13 adjectives with frequency >= 1000 in train
% [('new', 2420), ('other', 2190), ('own', 1500), ('such', 1380), ('more', 1359), ('many', 1348), ('first', 1220), ('great', 1206), ('same', 1116), ('different', 1069), ('good', 1006), ('political', 1005), ('local', 1002)]
% 5023 test phrases contain over limit adj from 47803, that's 10.5% of the test set

% 52 adjectives with frequency >= 500 in train
% [('new', 2420), ('other', 2190), ('own', 1500), ('such', 1380), ('more', 1359), ('many', 1348), ('first', 1220), ('great', 1206), ('same', 1116), ('different', 1069), ('good', 1006), ('political', 1005), ('local', 1002), ('major', 869), ('small', 795), ('current', 759), ('national', 731), ('particular', 729), ('public', 724), ('large', 714), ('real', 712), ('American', 697), ('social', 678), ('individual', 668), ('main', 661), ('little', 640), ('important', 629), ('international', 617), ('single', 617), ('economic', 617), ('old', 613), ('only', 608), ('specific', 607), ('various', 593), ('few', 592), ('personal', 591), ('general', 583), ('further', 569), ('human', 569), ('best', 561), ('big', 556), ('recent', 553), ('certain', 547), ('second', 539), ('full', 536), ('original', 533), ('several', 523), ('future', 518), ('special', 516), ('high', 508), ('most', 506), ('similar', 504)]
% 11849 test phrases contain over limit adj from 47803, that's 24.8% of the test set

The additive models (Addition, SAddition, VAddition) are the least competitive models in our evaluation, on all datasets. The results strongly argue for the point that \emph{additive models are too limited for composition}. An adequate composed representation cannot be obtained simply as an (weighted) average of the input components.
% Interestingly, however, the additive models obtain their best performance on the three English datasets (~50%, even more for adv-adj). Is it forms vs lemmas?

The Matrix model clearly outperforms the additive models. However, its results are modest in comparison to models like WMask+, BiLinear, FullLex+ and TransWeight. This is to be expected: having a single affine transformation limits the model's capacity to adapt to all the possible input vectors $\mathbf{u}$ and $\mathbf{v}$. Because of its small number of parameters, the Matrix model can only capture the general trends in the data.

More interaction between $\mathbf{u}$ and $\mathbf{v}$ is promoted by the BiLinear model through the $d$ bilinear forms in the tensor $\mathbf{E}\in\mathbb{R}^{n\times d\times n}$. This capacity to absorb more information from the training data translates into better results ---  the BiLinear model outperforms the Matrix model on all datasets.

In evaluating FullLex we tried to mitigate its treatment of unknown words. Instead of using unknown matrices to model composition of phrases not in the training data, we take a nearest neighbor approach to composition. Take for example the phrase \textit{sky-blue dress},  where \textit{sky-blue} does not occur in \texttt{train}. Our implementation, FullLex+, looks for the nearest neighbor of \textit{sky-blue} that appears in \texttt{train}, \textit{blue} and uses the matrix associated with it for building the composed representation. The same approach is also used for the WMask model, which is referred to as WMask+.

The use of datasets with a range of different sizes revealed that datasets with a smaller number of phrases per unique word can be successfully modelled using only transformation vectors. However, datasets with a larger number of phrases per word require the use of transformation matrices in order to generalize. For example, the Dutch compounds dataset has 5,317 unique words and 17,773 phrases, resulting in 3.3 phrases per word. On this dataset WMask+ fares only slightly worse than FullLex+ (0.70\%), an indication that FullLex+ suffers from data sparsity in such scenarios and cannot produce good results without an adequate amount of training data. By contrast, the gap between the two models increases considerably on datasets with more phrases per word --- e.g. FullLex+ outperforms WMask+ with 8.07\% on the English adjective-noun phrase dataset, which has 11.6 phrases per word.\footnote{The number of unique words and phrases for each dataset is available in Table~\ref{tbl:composition-datasets}.}

We compared FullLex+ and TransWeight in terms of their ability to
model phrases where at least one of the constituents is not part of
the training data. For example 16.2\% of the test portion of the
English compounds dataset, 563 compounds, have at least one
constituent that is not seen during training. We evaluated FullLex+
and TransWeight on this subset of data: 59.15\% of the representations
composed using FullLex+ obtain a rank $\leq5$. When using
TransWeight, a rank $\leq5$ is obtained for 67.50\% of the
representations. The difference between the two results is an
indicator of the superior generalization capabilities of TransWeight.

% (Corina) I tested this on EN adj-n first, there are 1684 phrases with one missing constituent, but that's only 3.5% of the test set, and there the results are closer - 73.63% for FullLex+ and 78.03% for TransWeight. 

TransWeight is the top performing composition model on small and large datasets alike. This shows that treating similar words similarly --- and not each word as a semantic island --- has a two-fold benefit: (i) it leads to good generalization capabilities when the training data is scarce and (ii) gives the model the possibility to accommodate a large number of training examples without increasing the number of parameters.

\subsection{Understanding TransWeight}
\label{sec:transweight_verstehen}

The results in Section~\ref{sec:comparison_results} have shown that
the transformations-based generalization strategy employed by
TransWeight works well across different languages and phrase
types. However, understanding what the transformations encode requires
taking a step back and contemplating again the architecture of the
model.

Each transformation used by the model can be seen as a separate
application of an affine transformation of the concatenated input
vectors $[\mathbf{u}; \mathbf{v}] \in \mathbb{R}^{2n}$ -- essentially,
one Matrix model -- resulting in a vector in $\mathbb{R}^n$. 100
transformations provide 100 ways of combining the same pair of input
vectors.

Two competing hypotheses can be put forth about the way each
transformation contributes to the final representation. The
\textit{specialization hypothesis} assumes that each transformation
specializes on particular input types e.g. bigrams made of color
adjectives and artifact-like nouns like \textit{black car}. In
contrast, the \textit{distribution hypothesis} assumes that the
parameters responsible for particular bigrams are distributed
across the transformations space instead of being confined to any
single transformation.

If the specialization hypothesis holds, removing the transformations
that are tailored to a particular input type will drastically reduce
the performance on instances of that input type. In order to test this
hypothesis, we evaluated TransWeight while randomly dropping full
transformations at dropout rates between 0\% and 90\% during
prediction.\footnote{The training hyperparameters are unchanged.} This
procedure also removes between 0\% and 90\% of the specialized
transformations --- assuming that they exist.

The performance of the model with transformation dropout is hard to
interpret in isolation, since it is to be expected that the
performance of a model decreases as side-effect of removing
parameters. Thus, any loss of performance can be attributed to the
removal of specific transformations \emph{or} to the reduction of the
number of parameters in general. In order to make the results
interpretable, we have created a reference model that drops individual
parameters of the transformed representations, rather than dropping
full transformations. The reference model removes the same number of
parameters as the model with transformation dropout, but keeps
specific transformations partially intact. This allows us to verify
whether the loss of performance of dropping out transformations is
larger than the expected loss of removing (any) parameters. If this is
indeed the case, removing certain transformations is more harmful than
removing random parameters and the specialization hypothesis should be
accepted. On the other hand, if there is no tangible difference
between the two models, then the specialization hypothesis should be
rejected in favor of the distribution hypothesis.

\begin{figure}[htbp]
  \centering
  \includegraphics[scale=0.5]{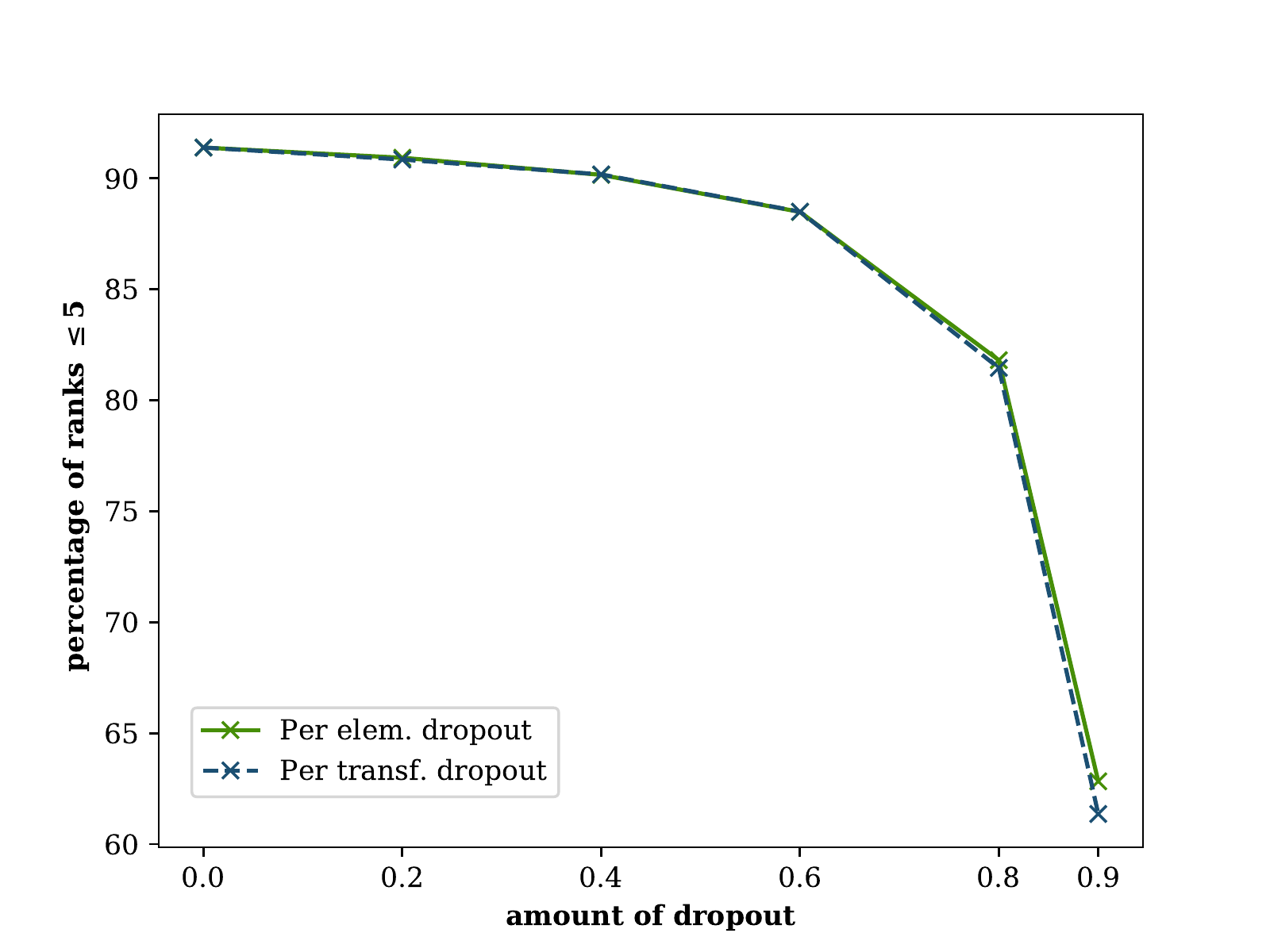}
  \caption{\label{fig:dropout_plot} The percentage of ranks $\leq 5$
    of the model that drops transformations randomly and the refence model
    that drops individual parameters of transformed representations
    randomly.}
\end{figure}

The results of this experiment on the English adjective-noun set are
shown in Figure~\ref{fig:dropout_plot}, which plots the percentage of
ranks $\leq 5$ against the dropout rate. Since there is virtually no
difference in losses between the model that uses transformation
dropout and the reference model, we reject the specialization
hypothesis. However, rejecting the specialization hypothesis does not
exclude the possibility that semantic properties of specific classes
of words are captured by parameters distributed across the
transformations.

%However, the individual
%dimensions of the vector space of $\mathbf{u}$, $\mathbf{v}$, and
%$\tilde{\mathbf{p}}$ encode semantic properties in terms of
%similarities between vectors, rather than individual vector components
%\cite{somepaper}. Moreover, the objective function does not add any
%semantic signal
%\cite{schulteimwalde,autoexendschutzerothe,hovy}. Therefore, we
%believe that semantic properties are also encoded in the similarity
%between transformed vectors and not in the transformations themselves.

% I (Daniël) decided to comment this out for now, seems to distact from
% the main point.
%
% It is also worthwhile to note the redundancy in the encoding of
% information in the transformations. Even when dropping out as much as
% 60\% of the transformations the percentage of test ranks $\leq5$,
% 88.50\%, remains comparable to the percentage obtained using the full
% set of transformations (91.39\%), and to the FullLex (90.74\%) and the
% BiLinear results (85.36\%) on the English adjective-noun dataset.
% This redundant encoding of information in the transformations is
% likely to be the result of training TransWeight using a high dropout
% rate of 0.8.

% \begin{table}[tbh!]
% \footnotesize
% \tabcolsep=0.04cm
% \centering

% \begin{tabular}{c|c|c?{0.4mm}c|c|c}
% \textbf{\makecell{Per elem.\\ dropout}} & \textbf{Quartiles} & 
% \textbf{$\leq$5} & \textbf{\makecell{Per transf.\\ dropout}} & \textbf{Quartiles} & 
% \textbf{$\leq$5}  \\
% \hline
% 0.0 & 1, 1, 2 & 91.39\% & 0.0 & 1, 1, 2 & 91.39\%\\
% 0.2 & 1, 1, 2 & 90.93\% & 0.2 & 1, 1, 2 & 90.85\%\\
% 0.4 & 1, 1, 2 & 90.16\% & 0.4 & 1, 1, 2 & 90.18\%\\
% 0.6 & 1, 1, 2 & 88.49\% & 0.6 & 1, 1, 2 & 88.50\%\\
% 0.8 & 1, 1, 4 & 81.81\% & 0.8 & 1, 1, 4 & 81.46\%\\
% 0.9 & 1, 3, 11 & 62.83\% & 0.9 & 1, 3, 12 & 61.36\%\\
% \hline
% \end{tabular}

% \caption{\label{tbl:tw_dropouts}}

% \end{table}

\section{Conclusion}

% Existing composition models are generally situated at two ends of a spectrum. They either have a small number of parameters but compose different words in the same way. Or they perform word-specific compositions at the cost of a large number of parameters. 

In this paper we have introduced TransWeight, a new composition model
which uses a set of weighted transformations, as a middle ground
between a fully lexicalized model and models based on a single
transformation. TransWeight outperforms all other models in our
experiments.

In this work, we have trained TransWeight for specific phrase
types. In the future, we would like to investigate whether a single
TransWeight model can be used to perform composition of different
phrase types, possibly while integrating information about the 
structure of the phrases and their context as in \citet{hermann2013role,yu2014factor,yu2015learning}.

Another extension that we are interested in is to use
TransWeight to compose more than two words. We plan to follow the lead
of \citet{socher2012semantic} here, which uses the FullLex composition
function in a recursive neural network to compose an arbitrary number
of words. Similarly, we could use TransWeight in a recursive neural
network in order to compose more than two words.

% Another interesting question that we hope to address is how many
% transformations are needed for different phrase types. 

In our experiments, 100 transformations yielded optimal results for
all phrase sets.  However, further investigation is needed to
determine if this number is optimal for any combination of word
classes, or whether it is dependent on the word class type (i.e. open
or closed), the diversity of the word classes in a dataset, or
properties of the embedding space that are inherent to the method used
to construct the vector space.

% The future might belong to a cross between FullLex and TransSel, that uses frequency information to
% decide between using a word-specific transformation and a category transformation

\section*{Acknowledgements}
We would like to thank our reviewers and in particular our action editor, Sebastian Pad\'{o},
for their constructive comments. We also want to thank the other members A3 project team for all their comments and suggestions during project meetings. Financial support for the research reported in this paper was provided by the German Research Foundation (DFG) as part of the Collaborative Research Center ``The Construction of Meaning'' (SFB 833), project A3.

\bibliography{tacl2018submission}
\bibliographystyle{acl_natbib}

\appendix

\begin{table*}[htb]
\footnotesize
\tabcolsep=0.14cm

\begin{tabular}{l?{0.4mm}r|rrr|c?{0.4mm}r|rrr|c?{0.4mm}r|rrr|c}
& \multicolumn{5}{c?{0.4mm}}{\textbf{Nominal Compounds}} & \multicolumn{5}{c?{0.4mm}}{\textbf{Adjective-Noun Phrases}} & \multicolumn{5}{c}{\textbf{Adverb-Adjective Phrases}} \\
\hline
\textbf{Model} & \textbf{Cos-d} & \textbf{Q$_1$} & \textbf{Q$_2$} & \textbf{Q$_3$} & \textbf{$\leq$5}
& \textbf{Cos-d} & \textbf{Q$_1$} & \textbf{Q$_2$} & \textbf{Q$_3$} & \textbf{$\leq$5}
& \textbf{Cos-d} & \textbf{Q$_1$} & \textbf{Q$_2$} & \textbf{Q$_3$} & \textbf{$\leq$5}\\
\hline

\hline
& \multicolumn{15}{c}{\textbf{English}} \\
\hline

WMask+ & 0.344 & 3 & 9 & 43 & 39.84\%
& 0.342 & 4 & 11 & 41 & 33.23\%
& 0.335 & 3 & 8 & 33 & 40.86\% \\
BiLinear & 0.335 & 3 & 7 & 33 & 43.81\% 
& 0.332 & 3 & 9 & 33 & 39.14\%
& 0.331 & 2 & 7 & 25 & 45.84\%\\ 
FullLex+ & 0.338 & 3 & 9 & 41.5 & 40.53\%
& 0.309 & \textbf{2} & \textbf{6} & \textbf{20} & \textbf{48.41\%}
& 0.327 & 2& 7 & 25 & 45.45\%\\
\textbf{TransWeight} & \textbf{0.30} & \textbf{2} & \textbf{7} & \textbf{34} & \textbf{44.44\%} 
& \textbf{0.307} & 3& 9 & 33 & 39.36\%
& \textbf{0.31} & \textbf{2} & \textbf{6} & \textbf{19} & \textbf{49.05\%} \\

\hline
& \multicolumn{15}{c}{\textbf{German}} \\
\hline

WMask+ &0.34 & 3 & 12 & 89 & 37.01\%
& 0.35 & 5 & 19 & 92 & 25.52\%
& 0.387 & 8 & 40 & 219.5 & 18.32\% \\ 
BiLinear & 0.334 & 3 & 11 & 97 & 38.73\%
& 0.34 & 5 & 15 & 74 & 29.51\%
& 0.383 & 7 & 29 & 163 & 21.83\% \\ 
FullLex+ & 0.328 & 3 & 10 & 72 & 39.89\% 
& 0.343 & \textbf{4} & \textbf{12} & \textbf{58}	& \textbf{32.49\%}
& 0.383 & 7 & 35 & 189 & 20.55\%\\
\textbf{TransWeight} & \textbf{0.31} & \textbf{2} & \textbf{10} & \textbf{76} & \textbf{40.10\%}  
& \textbf{0.324} & 4 & 14 & 68 & 30.24\%
& \textbf{0.367} & \textbf{6} & \textbf{24} & \textbf{130} & \textbf{23.21\%} \\

\hline
& \multicolumn{15}{c}{\textbf{Dutch}} \\
\hline

WMask+ & 0.393 & 7 & 39 & 307 & 21.09\% 
& 0.378 & 6 & 20 & 132 & 24.87\% 
& 0.429 & 9 & 44 & 315 & 17.42\%\\ 
BiLinear & 0.396 & 6& 34 & 284 & 24.37\% 
& 0.375 & 5 & 19 & 140 & 27.40\% 
& 0.426 & 8 & 34 & 239 & \textbf{19.18\%} \\ 
FullLex+ & 0.388 & 6 & 37 & 313.5 & 22.93\%
& 0.362 & \textbf{4} & \textbf{13} & \textbf{80} &\textbf{31.57\%} 
& 0.433 & 10 & 50 & 381 & 16.65\% \\
\textbf{TransWeight} & \textbf{0.376} & \textbf{5} & \textbf{28} & \textbf{235.5} & \textbf{25.25\%} 
& \textbf{0.349} & 5 & 18 & 16 & 26.92\%
& \textbf{0.423} & \textbf{8} & \textbf{29} & \textbf{238} & 18.96\% \\

\hline
\end{tabular}

\caption{\label{tbl:results_oldEval} Results using the \citet{baroni2010nouns} evaluation method for the best performing models for English, German and Dutch on the composition of nominal compounds, adjective-noun phrases and adverb-adjective phrases.}
\end{table*}

\FloatBarrier

\section{Hyperparameters word embeddings}
\label{appendix:embed-hyperparameters}

The word embeddings were trained using the skip-gram model with
negative sampling \cite{mikolov2013distributed} from the
\texttt{word2vec} package. Arguments: embedding size of 200; symmetric
window of size 10; 25 negative samples per positive training instance;
and a sample probability threshold of $10^{-4}$. As a default, we use
a minimum frequency cut-off of 50. However, for German
adverb-adjective phrases and all Dutch phrases we used a cut-off to 30
to be able to extract enough phrases for training and evaluation.

\section{Hyperparameters composition models}
\label{appendix:hyperparameters}

Dropout \cite{srivastava2014dropout} rates between 0 and 0.8 in 0.2 increments were tested on the \texttt{dev} set for the four weighting variations presented in Table~\ref{tbl:weighting_variations}, while keeping constant the number of transformations (100). TransWeight-r performed best with a dropout of 0.4, TransWeight-c and TransWeight-v with 0.6 dropout and TransWeight with 0.8 dropout. For TransWeight the dropout is applied to $\mathbf{H}$, the matrix containing the transformed representations.

% \section{\citet{baroni2010nouns} evaluation}
% \label{appendix:baroni-eval}

% For TransWeight we experimented with using between 20 to 500 transformations. Using less than 80 transformations resulted in suboptimal performance on all datasets. However, increasing the number of transformations above 100 only provided diminishing results at the cost of a larger number of parameters. 

% For a better comparison all the TransWeight models in Tables~\ref{tbl:results_compounds}, \ref{tbl:results_adj_noun_sample_word2vec} and \ref{tbl:results_adj_noun_full_word2vec}, as well as the variations in Table~\ref{tbl:weighting_variations} use 100 transformations.

% ? normalization, implementation details (dropout, learning rate), high (bad) rank instances

\end{document}